\documentclass[11pt]{article}

\usepackage[preprint]{acl}
\usepackage{bandai_acl}
\usepackage{times}
\usepackage{latexsym}

\usepackage[T1]{fontenc}

\usepackage[utf8]{inputenc}

\usepackage{microtype}

\usepackage{inconsolata}

\usepackage{graphicx}

\usepackage{fancyhdr}
\usepackage{url}
\usepackage{tcolorbox}
\usepackage{wrapfig}
\usepackage{hyperref}       
\usepackage{url}            
\usepackage{booktabs}       
\usepackage{amsfonts}       
\usepackage{nicefrac}       
\usepackage{microtype}      
\usepackage{xcolor}         
\usepackage{amsmath,amssymb}
\usepackage{algorithmic}
\usepackage{multirow}
\usepackage{xcolor}
\usepackage{tcolorbox}
\tcbuselibrary{breakable}
\usepackage{paracol}
\usepackage[utf8]{inputenc} 
\usepackage{CJKutf8}
\usepackage{tcolorbox}
\usepackage[table]{xcolor}
\tcbuselibrary{breakable,skins}

\fancypagestyle{firstpagestyle}{
  \fancyhf{}
  \fancyhead[L]{\includegraphics[width=70pt]{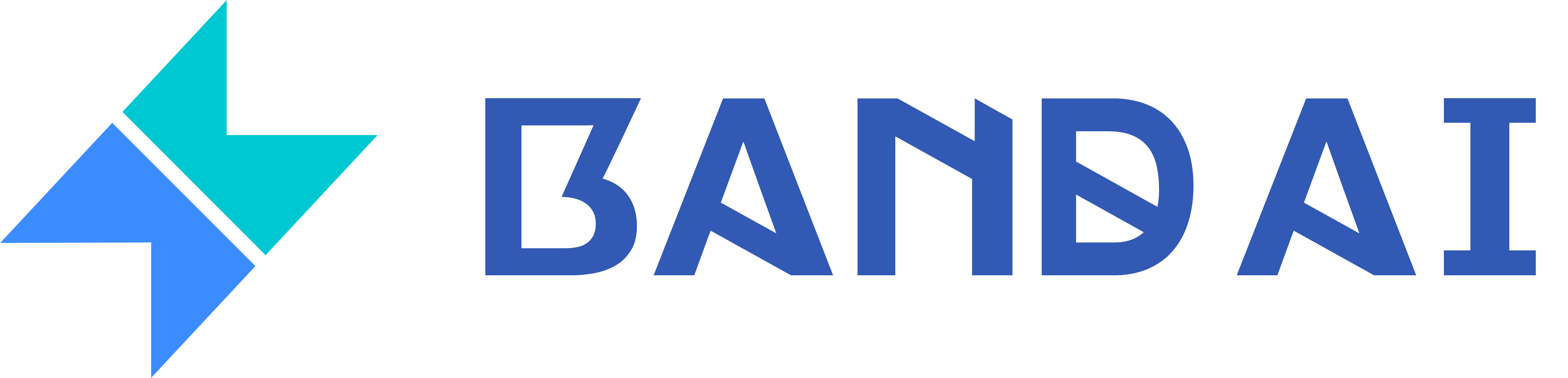}}
}

\fancypagestyle{plain}{
  \fancyhf{}     
}

\newtcolorbox{dialoguebox}[1][]{
  breakable,
  colback=gray!3,
  colframe=gray!40,
  boxrule=0.5pt,
  arc=2pt,
  left=6pt,
  right=6pt,
  top=6pt,
  bottom=6pt,
  fontupper=\small,
  colbacktitle=gray!25,    
  coltitle=black,
  title=#1,
  fonttitle=\small\bfseries,
}

\newcommand{\zh}[1]{%
\begin{CJK}{UTF8}{gbsn}#1\end{CJK}
}

\newcommand{\en}[1]{%
\par\small\textit{#1}\par
}

\title{TeachBench: A Syllabus-Grounded Framework for Evaluating Teaching Ability in Large Language Models}

\author{
Zheng Li$^{1,2}$,
Siyao Song$^{2,3}$,
Jingyuan Ma$^{1,2}$,
Rui Li$^{1,2}$,
Ying Zeng$^{2}$,
Minghao Li$^{2}$\thanks{Corresponding authors.},
Zhifang Sui$^{1}$\footnotemark[\value{footnote}] \\
\\
$^{1}$State Key Laboratory of Multimedia Information Processing, School of Computer Science, \\Peking University 
$^{2}$ByteDance BandAI\symbolbandai, 
$^{3}$Institute of Automation, Chinese Academy of Sciences \\
\small\texttt{lizheng2001@pku.edu.cn; songsiyao2024@ia.ac.cn; minghao.li@bytedance.com}
}

\begin{document}
\thispagestyle{firstpagestyle}   
\pagestyle{plain}    
\maketitle
\begin{abstract}
Large language models (LLMs) show promise as teaching assistants, yet their teaching capability remains insufficiently evaluated. Existing benchmarks mainly focus on problem-solving or problem-level guidance, leaving knowledge-centered teaching underexplored. We propose a syllabus-grounded evaluation framework that measures LLM teaching capability via student performance improvement after multi-turn instruction. By restricting teacher agents to structured knowledge points and example problems, the framework avoids information leakage and enables reuse of existing benchmarks. We instantiate the framework on Gaokao data across multiple subjects. Experiments reveal substantial variation in teaching effectiveness across models and domains: some models perform well in mathematics, while teaching remains challenging in physics and chemistry. We also find that incorporating example problems does not necessarily improve teaching, as models often shift toward example-specific error correction. Overall, our results highlight teaching ability as a distinct and measurable dimension of LLM behavior.
\end{abstract}

\section{Introduction}
\label{introduction}
Large language models (LLMs) have demonstrated strong performance across a wide range of reasoning and problem-solving tasks. These advances have made LLMs promising candidates for AI-powered teaching assistants and automated educational systems. Recent work has begun to explore the use of LLMs in educational settings, such as interactive tutoring and learning support\cite{nye2023generative,liu2024socraticlm,patil2024automated}. However, despite increasing interest in these applications, systematic benchmarks for evaluating the teaching capability of LLMs remain largely underdeveloped.

Several recent benchmarks have begun to explore educational scenarios. For example, datasets such as EducationQ\cite{shi-etal-2025-educationq} and Teach2Eval\cite{zhou2025teach2eval} evaluate whether an LLM can guide a student toward solving a specific target problem. While valuable, these benchmarks primarily focus on problem-level guidance rather than teaching ability in a broader sense. In particular, whether a model can effectively act as a teacher—helping students learn underlying knowledge points and acquire problem-solving strategies—remains insufficiently evaluated. Moreover, directly teaching target questions introduces the risk of information leakage, which can invalidate evaluation results.

In this work, we propose a teaching-centered evaluation framework that explicitly measures the instructional effectiveness of LLMs acting as teacher agents. Our framework is built around structured knowledge points derived from a syllabus, rather than target questions themselves. A teacher agent conducts multi-turn instruction based on these knowledge points, while a fixed-capability student agent serves as a proxy for a human learner. Teaching effectiveness is assessed by the student's performance improvement after instruction, enabling controlled and fair comparison across different teacher agents.

To support this evaluation, we construct a knowledge structure tree from exam syllabi and annotate each question with fine-grained knowledge paths. We further generate example problems of varying difficulty for each knowledge point, allowing us to study different teaching strategies under a unified protocol. Using data from the Chinese National College Entrance Examination (Gaokao) across multiple subjects, we conduct extensive experiments to evaluate teaching effectiveness under different interaction settings.

Our experimental results reveal several key findings. First, among the evaluated models, Qwen3-235B-A22B-Instruct demonstrates strong teaching effectiveness, achieving a 7.63-point improvement in the mathematics domain. Second, teaching effectiveness varies substantially across domains: models perform better in subjects where knowledge points can be directly applied (e.g., mathematics, history, and politics), while teaching remains more challenging in domains requiring deeper integration of knowledge with complex problem contexts (e.g., physics and chemistry). Finally, we find that incorporating example problems does not necessarily enhance teaching effectiveness. Instead, current LLMs tend to shift from syllabus-grounded instruction to example-based error correction, which fragments the teaching process and ultimately weakens overall instructional quality.

Together, these results highlight both the potential and limitations of current LLMs as teachers. More importantly, they show that teaching ability is a distinct and measurable dimension of LLM behavior, separate from raw problem-solving performance. We hope our framework provides a foundation for future research on LLM pedagogy and the development of models that can teach more effectively.

Our main contributions are summarized as follows:
\begin{itemize}
    \item We propose a syllabus-grounded evaluation framework that measures LLM teaching capability via post-instruction student performance, while preventing information leakage by restricting teachers to structured knowledge points and enabling reuse of existing benchmarks.
    \item We construct a knowledge-structured benchmark based on Gaokao syllabi and questions, including fine-grained knowledge annotation and controlled Student–Teacher interactions across multiple subjects.
    \item Through extensive experiments, we provide a systematic analysis of LLM teaching behaviors, revealing both effective teaching patterns and key limitations, particularly in example-based instruction.
\end{itemize}

\section{Related Work}
\label{relatedwork}
\paragraph{Answer-centric benchmarks for knowledge and reasoning.}
Mainstream LLM evaluation has predominantly focused on \emph{answer correctness} under static test settings, measuring knowledge coverage and reasoning skill rather than instructional competence. Representative suites include MMLU for broad academic/professional knowledge \cite{hendrycks2021mmlu}, HELM for standardized multi-scenario and multi-metric reporting \cite{liang2022helm}, and math-focused benchmarks such as GSM8K and MathBench for multi-step quantitative reasoning and hierarchical proficiency assessment \cite{cobbe2021gsm8k,liu-etal-2024-mathbench}. While these benchmarks are invaluable for characterizing “solver” capability, they do not operationalize whether a model can act as a teacher who improves a learner’s mastery.
\emph{TechBench addresses this gap} by shifting the evaluation target from solving to teaching: the model is assessed by its ability to organize instruction around a syllabus-derived knowledge structure and to induce measurable learning gains under controlled instructional inputs.

\paragraph{Real-exam and Chinese-context evaluation.}
To increase ecological validity, several benchmarks evaluate LLMs using standardized exams and Chinese-context tests. AGIEval adopts a human-exam-centric design spanning multiple official exams \cite{zhong-etal-2024-agieval}; C-Eval and CMMLU provide large-scale Chinese multi-discipline multiple-choice suites across difficulty levels \cite{huang2023ceval,li-etal-2024-cmmlu}; and GAOKAO-Bench directly evaluates LLM performance on Gaokao questions, including subjective formats with grading considerations \cite{zhang2023gaokao}. Despite their realism, these efforts remain largely \emph{solution-centric}: they quantify how well a model answers exam questions, but not how well it can teach the underlying knowledge without being shown the test item.
\emph{TechBench differs} by explicitly instantiating a Gaokao-like \textbf{teaching} scenario: the teacher agent is not given the target question and must teach solely from knowledge points plus curated example problems, and success is defined by post-instruction improvement on held-out tests rather than raw exam scoring alone.

\begin{figure*}[htbp]
    \centering
    \includegraphics[width=\linewidth]{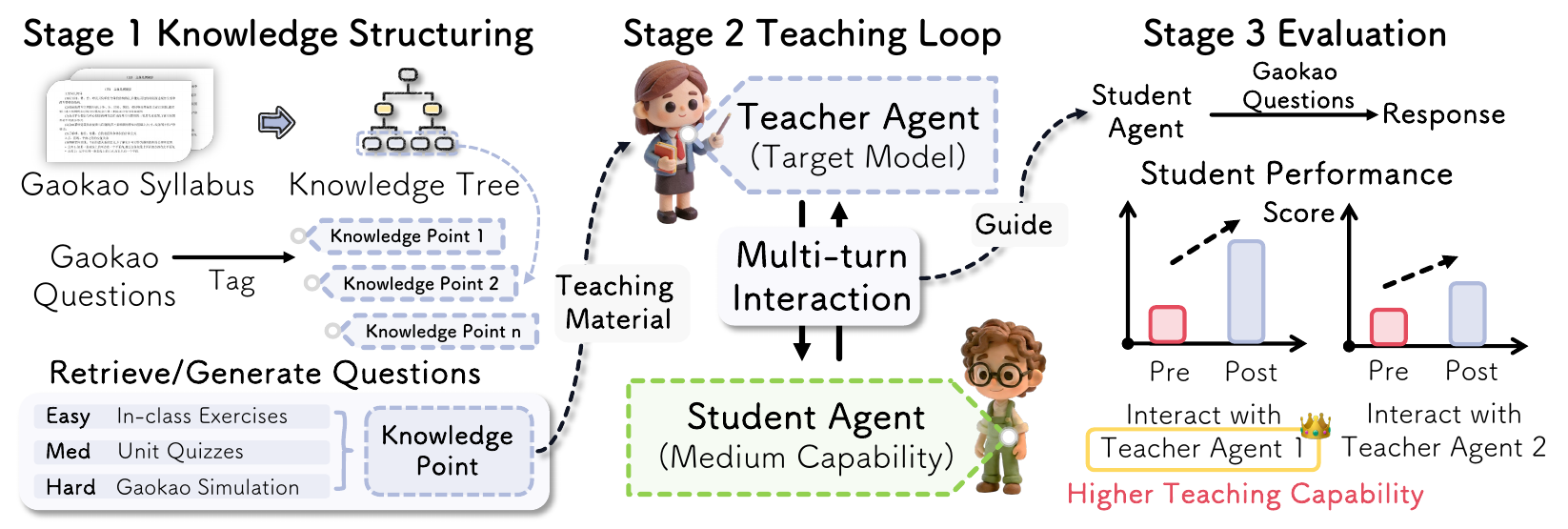}
    \caption{Overview of the proposed LLM teaching evaluation framework. The workflow consists of three stages: (1) \textbf{Knowledge Structuring}, which transforms the raw syllabus and questions into a structured knowledge tree with associated exercises; (2) \textbf{Teaching Loop}, which simulates multi-turn, dialogue-based interactions and instruction between the target LLM acting as the teacher agent and a student agent; and (3) \textbf{Evaluation}, which quantifies student performance improvement by comparing results before and after the interaction, yielding a final teaching ability score.}
    \label{fig:teaser}
\end{figure*}

\paragraph{Tutoring datasets and pedagogy-oriented benchmarks.}
Recent work has begun to formalize pedagogical behaviors through tutoring dialogues and rubric-based evaluation. MathDial provides teacher--student tutoring dialogues grounded in multi-step math problems with pedagogical properties \cite{macina-etal-2023-mathdial}. The BEA shared tasks benchmark the generation and assessment of teacher responses in educational dialogues, highlighting both progress and the limitations of generic automatic metrics for educational contexts \cite{tack-etal-2023-bea,kochmar-etal-2025-bea}. Beyond shared tasks, TutorBench and MathTutorBench evaluate tutoring skills such as adaptive explanations, feedback, and hinting, often via rubric-based or learned scoring \cite{srinivasa2025tutorbench,macina-etal-2025-mathtutorbench,shi-etal-2025-educationq,zhou2025teach2eval,lelievre2025benchmarking,liu2024socraticlm,maurya2025unifying}.
\emph{TeachBench complements these lines} by targeting a different notion of pedagogical success: rather than primarily scoring per-turn response quality, it evaluates \textbf{end-to-end teaching effectiveness} under a knowledge-tree curriculum and leakage-controlled inputs.

\section{Method}
\label{method}
To evaluate the teaching capability of LLMs when acting as genuine teachers, we design a workflow that leverages a knowledge syllabus and corresponding exam questions to assess an LLM's ability to teach specific knowledge points. As illustrated in Figure~\ref{fig:teaser}, our approach first extracts the syllabus into a structured knowledge tree using a semi-automatic pipeline. Based on this knowledge tree, an LLM-based Tagger is applied to annotate the knowledge points involved in each question. In parallel, an LLM-based Question Generator produces example problems of varying difficulty levels for each knowledge point.

Finally, we employ an LLM with moderate capability as a student agent and conduct multi-turn interactions between the student agent and the teacher agent under evaluation. The teaching ability of the teacher agent is assessed through the improvement in the student's performance over the course of the interaction.

\subsection{Knowledge Structure Tree}

To facilitate downstream processing, we convert the syllabus of exam topics into a structured tree representation. We first use Gemini-3\cite{deepmind_gemini_models} to process the syllabus and extract an initial knowledge structure tree. We then employ GPT-5\cite{openai_gpt5} to review the tree and supplement any missing knowledge points. Finally, the resulting knowledge structure tree is manually inspected to ensure its accuracy and completeness, such that it fully covers the intended scope of the syllabus.

\subsection{LLM-based Question Tagger}

Each question may involve multiple knowledge points. Given a question and a set of candidate knowledge nodes at the current level of the knowledge hierarchy, we prompt an LLM to identify all knowledge nodes that are relevant to the question. For each selected node, the tagging process is then recursively applied to its child nodes.

We adopt a depth-first traversal strategy over the knowledge hierarchy. All paths expanded from the selected nodes down to leaf nodes are collected, and each root-to-leaf path is treated as a knowledge tag associated with the question. The prompt template used for question tagging is shown in Appendix~\ref{appendix: prompt templates}.

\subsection{LLM-based Question Generator}
To provide the teacher agent with pedagogically appropriate practice material while avoiding direct exposure to target exam items, we construct an LLM-based Question Generator that synthesizes (or retrieves) representative example problems for each knowledge point in the knowledge structure tree. Concretely, given a leaf-level knowledge point $k$, together with its full path in the hierarchy to disambiguate scope, the generator outputs a small set of scaffolded exercises with ground-truth answers and step-by-step solutions. These examples serve as the only task-specific materials available to the teacher agent during instruction (cf. \S\ref{sec:teacher-agent}), thereby reducing the risk of information leakage from target questions and enabling a controlled evaluation of teaching effectiveness.

\paragraph{Model choice and web-enabled retrieval.}
We use a web-enabled strong LLM, Gemini-2.5-Pro, to support both retrieval and generation. For each knowledge point $k$, we first prompt the model to issue web queries that combine the canonical name of $k$ and the subject context implied by the syllabus. If the model can retrieve high-quality and on-topic example problems from the web, we prioritize these retrieved items as they better reflect naturally occurring instructional materials. For each retrieved candidate, we ask the model to extract a clean problem statement, standardize notation, and reconstruct a complete answer and solution in a unified format. When no suitable web results are found (or retrieved candidates are low-quality, off-topic, or incomplete), the model is instructed to generate new problems from scratch under the same formatting and pedagogical constraints.

\paragraph{Difficulty calibration and output specification.}
For each knowledge point $k$, the generator produces three problems with an explicit difficulty gradient:
(1) \textbf{Level 1 (easy)} targets basic comprehension and direct application, akin to in-class exercises;
(2) \textbf{Level 2 (medium)} targets after-class practice and requires non-trivial reasoning or multi-step computation, akin to unit quizzes;
(3) \textbf{Level 3 (hard)} targets integrated and challenging reasoning comparable to the overall difficulty profile of the National College Entrance Examination (Gaokao), but \emph{must not} directly reuse any official Gaokao items.
To enforce this constraint, we instruct the generator to avoid verbatim reuse of known past exam problems and instead create isomorphic or novel variants while preserving alignment with $k$. Each generated (or retrieved and normalized) item is serialized as a structured record, facilitating downstream prompting of the teacher agent and supporting automated auditing.

\paragraph{Post-generation validation and self-consistency checks.}
Because small factual or logical errors in example problems can confound teaching evaluation, we apply a second-pass verification stage, also implemented with Gemini-2.5-Pro. The verifier is prompted to (i) check knowledge-point alignment, (ii) verify answer correctness and solution validity, and (iii) assess \textbf{difficulty appropriateness} relative to the Level 1--3 rubric. If any check fails, the item is either revised or regenerated. We repeat this repair/regeneration until the verifier accepts the item under all criteria.

\begin{table*}[t]
\centering
\begin{tabular}{ccccccccc}
\toprule
\textbf{Subject} & Math & Physics & Chemistry & Biology & History & Geography & Politics & Total\\
\midrule
\textbf{Num} & 166 & 143 & 140 & 165 & 125 & 175 & 175 & 1,089\\
\bottomrule
\end{tabular}
\caption{Distribution of evaluation questions across subjects.}
\label{tab:dataset static}
\end{table*}

\subsection{Teacher Agent}
\label{sec:teacher-agent}
To evaluate the teaching capability of LLMs in the role of a teacher, while preventing potential information leakage that may arise from directly teaching the target questions, we restrict the inputs provided to the Teacher LLM. Specifically, the Teacher LLM is given only the knowledge points associated with each question, along with corresponding example problems, rather than the target question itself. The Teacher LLM is then instructed to conduct instruction based solely on these knowledge points.

The teacher LLM is instructed to teach the specified knowledge points in sequence and to assess the student's understanding at each step, determining when to advance or terminate the instruction (see Appendix~\ref{appendix: prompt templates}).

\subsection{Student Agent}

Given the strong reasoning capabilities of contemporary LLMs, we use an LLM as a proxy for a human student. The system prompt of the Student LLM specifies the target knowledge points to be learned and instructs the Student LLM to behave as a learner acquiring new knowledge (see Appendix~\ref{appendix: prompt templates} for details). During the interaction, the Student LLM is required to cooperate with the teacher agent and engage in the learning process centered around the given knowledge points.

\subsection{Teaching Procedure}

The teacher agent and the student agent interact through a dialogue-based teaching process. For each knowledge point, the teacher agent first provides an initial explanation, to which the student agent responds based on its current understanding. The teacher agent then evaluates the student's response to determine whether the knowledge point has been sufficiently mastered, and accordingly decides whether to continue explaining the same knowledge point or to proceed to the next one.

The teaching process continues in this manner until the teacher agent determines, based on the student agent's responses, that all target knowledge points have been adequately mastered. At this stage, the teacher agent terminates the teaching procedure.

\subsection{Evaluation of Teaching Effectiveness}

To evaluate the teacher agent's ability to teach specific knowledge points, we measure the improvement in the student agent's performance before and after instruction. Prior to the teaching process, the student agent is asked to directly answer the target questions, providing an estimate of its initial capability.

The teaching dialogue is then initiated from scratch, during which the teacher agent conducts instruction based on the associated knowledge points. After the teaching process is completed, the student agent, with access to the teaching dialogue history, is asked to answer the corresponding questions again to assess its post-instruction performance.

The improvement in accuracy between the pre-instruction and post-instruction evaluations reflects the change in the student's capability and is used as the primary metric for assessing the teacher agent's teaching effectiveness.

\section{Experiment}
\label{experiment}

\subsection{Experiment Data}


Given that the Chinese National College Entrance Examination (Gaokao) is accompanied by a well-defined knowledge syllabus and a large collection of exam questions, we adopt Gaokao syllabi and questions as our experimental data. We collect knowledge point syllabi for seven subjects (Mathematics, Physics, Chemistry, Biology, History, Geography, and Politics) from publicly available online sources, and use the corresponding Gaokao questions from C-Eval as the test set.

For each question, we attempt the knowledge tagging process up to six times to mitigate failures caused by network instability. We retain only questions for which tagging is successful. The final dataset contains 1,089 questions in total, with the number of questions per subject reported in Table~\ref{tab:dataset static}.

\subsection{Experiment Settings}

\begin{figure}[t]
    \centering
    \includegraphics[width=\linewidth]{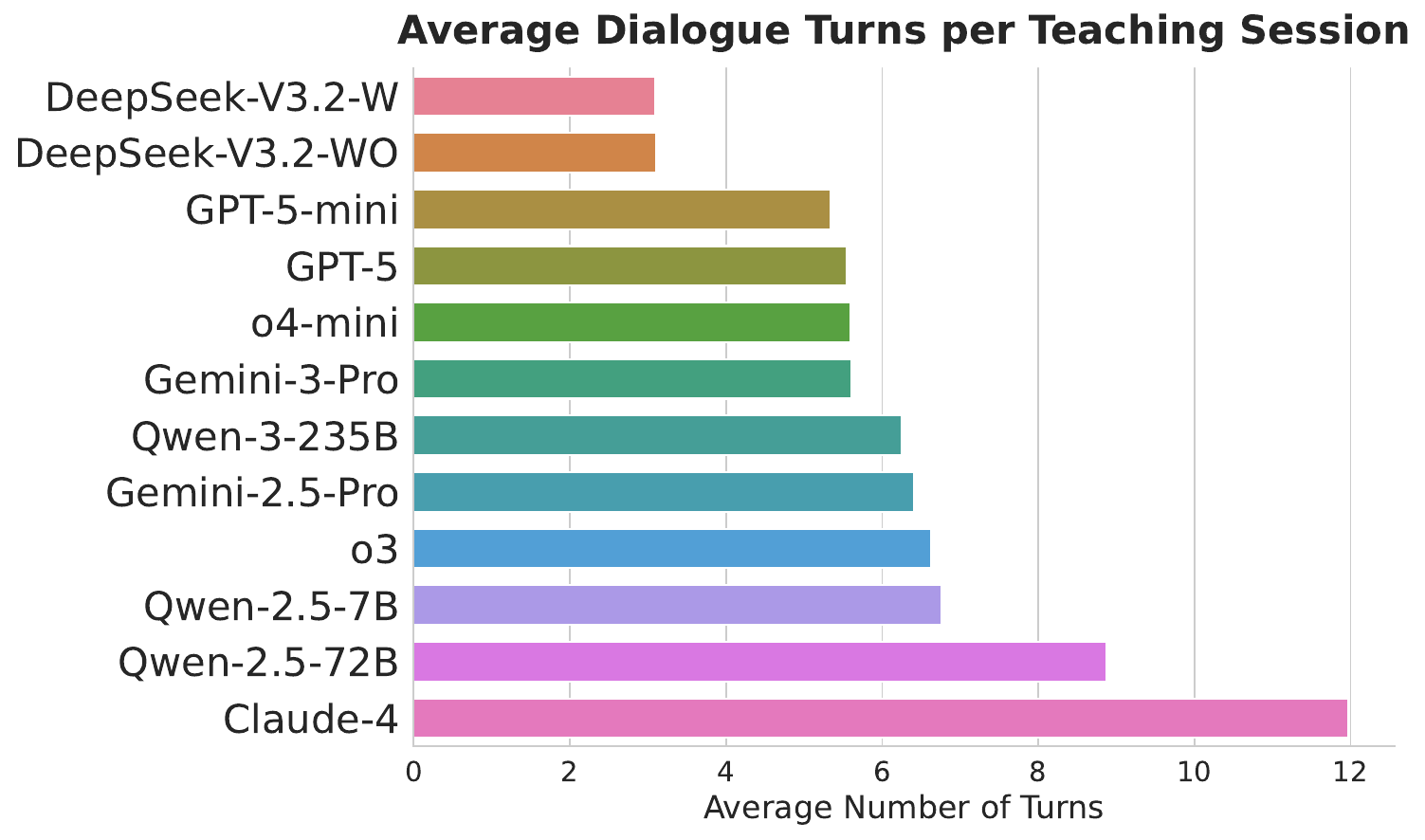}
    \caption{Average number of dialogue turns required by different models to complete a teaching session.}
    \label{fig:turns-dia}
\end{figure}

During the experiments, we aim for the student agent to possess sufficient learning capability while still leaving room for performance improvement on the test data. To this end, we select Qwen2.5-7B-Instruct as the student agent. We evaluate a range of mainstream open-source and commercial LLMs as teacher agents. 

\begin{table*}[t]
\centering
\begin{tabular}{lrrrrrrrr}
\toprule
\multirow{2}{*}{Models} & \multicolumn{2}{c}{Pass@1} & \multicolumn{2}{c}{Pass@4} & \multicolumn{2}{c}{Pass@16} & \multicolumn{2}{c}{Pass@64} \\
\cmidrule(lr){2-3} \cmidrule(lr){4-5} \cmidrule(lr){6-7} \cmidrule(lr){8-9} 
 & Acc & $\Delta$ & Acc & $\Delta$ & Acc & $\Delta$ & Acc & $\Delta$ \\
\midrule
\rowcolor{gray!15}
Qwen2.5-7B-Instruct & 60.53  & -  & 87.35  & -  & 94.58 & - & 97.59  & -  \\
\rowcolor{gray!15}
Qwen2.5-7B-Instruct(Knowledge) & 64.90  & 4.37 & 84.74  & -2.61 & 94.38 & -0.20 & 96.99  & -0.60   \\
\midrule 
Qwen2.5-7B-Instruct & 66.54  & 6.01  & 85.54  & -1.81  & 95.38 & 0.80 & 98.19  & 0.60  \\
Qwen2.5-72B-Instruct & 67.27 & 6.74 & 84.54 & -2.81 & 94.78 & 0.20 & 97.79 & 0.20 \\
Qwen3-235B-A22B-Instruct & 68.16 & 7.63 & 87.35 & 0.00 & 94.58 & 0.00 & 97.59 & 0.00 \\
GPT-5 & 67.63 & 7.10 & 85.54 & -1.81 & 94.98 & 0.40 & 98.19 & 0.60 \\
GPT-5-mini & 67.63 & 7.10 & 87.75 & 0.40 & 96.18 & 1.60 & 97.59 & 0.00 \\
OpenAI o3 & 67.69 & 7.16 & 87.35 & 0.00 & 94.78 & 0.20 & 97.59 & 0.00 \\
OpenAI o4-mini & 67.57 & 7.04 & 84.14 & -3.21 & 94.78 & 0.20 & 97.79 & 0.20 \\
Gemini-3-Pro & 67.71 & 7.18 & 86.35 & -1.00 & 95.18 & 0.60 & 98.19 & 0.60 \\
Gemini-2.5-Pro & 68.13 & 7.60 & 86.75 & -0.60 & 95.38 & -0.20 & 97.99 & 0.40 \\
Claude-4-Opus & 67.65 & 7.12 & 84.11 & -3.24 & 93.82 & 0.76 & 97.79 & 0.20 \\
DeepSeek-V3.2-w/ Thinking & 66.54 & 6.01 & 85.54 & -1.81 & 95.18 & 0.60 & 97.39 & -0.20 \\
DeepSeek-V3.2-w/o Thinking & 67.39 & 6.86 & 86.35 & -1.00 & 95.18 & 0.60 & 97.39 & -0.20 \\
\bottomrule
\end{tabular}
\caption{Results of teaching-based evaluation on the Gaokao Mathematics subset. We report student performance under different teacher agents across varying sampling budgets (Pass@k). ``Acc'' denotes absolute accuracy, and ``$\Delta$'' indicates the improvement over the no-teaching baseline. The upper part of the table reports baselines where the student model answers questions directly or with access to knowledge points only, while the lower part shows results after instruction by different Teacher LLMs.} 
\label{tab:main results performance}
\end{table*}

During the teaching dialogue, the termination of instruction is determined by monitoring a predefined end-of-teaching token, ``Teach Done'', generated by the teacher agent. We further observe that, without an upper bound on the total number of dialogue turns, some models may fail to emit the termination token after completing instruction and instead enter repetitive exchanges (e.g., mutual acknowledgments) with the student agent. We analyze the average number of dialogue turns during teaching for different models, as shown in Figure~\ref{fig:turns-dia}. We observe that teaching interactions typically conclude within 30 turns across models. To prevent potential dead loops between the teacher agent and the student agent, we therefore set the maximum number of dialogue turns to 30.

To ensure the stability of experimental results, each teacher agent conducts three independent teaching sessions. For each teaching session, the student agent performs 64 repeated evaluations using the corresponding teaching dialogue history. We report both the average performance and the best-of-N (BoN) results of the student agent.


\begin{table*}[t]
\centering
\begin{tabular}{lrrrrrrrr}
\toprule
\multirow{2}{*}{Models} & \multicolumn{2}{c}{Pass@1} & \multicolumn{2}{c}{Pass@4} & \multicolumn{2}{c}{Pass@16} & \multicolumn{2}{c}{Pass@64} \\
\cmidrule(lr){2-3} \cmidrule(lr){4-5} \cmidrule(lr){6-7} \cmidrule(lr){8-9}
 & Acc & $\Delta$ & Acc & $\Delta$ & Acc & $\Delta$ & Acc & $\Delta$ \\
\midrule
Qwen2.5-7B-Instruct & 60.53  & -  & 87.35  & -  & 94.58 & - & 97.59  & -  \\
Qwen2.5-7B-Instruct(Knowledge) & 64.90  & 4.37 & 84.74  & -2.61 & 94.38 & -0.20 & 96.99  & -0.60   \\
\midrule 
GPT-5 & 67.63 & 7.10 & 85.54 & -1.81 & 94.98 & 0.40 & 98.19 & 0.60 \\
GPT-5-Example & 64.59 & 4.06 & 86.36 & -0.99 & 94.81 & 0.23 & 98.05 & 0.46 \\

Gemini-2.5-Pro & 68.13 & 7.60 & 86.75 & -0.60 & 95.38 & -0.20 & 97.99 & 0.40 \\
Gemini-2.5-Pro-Example & 63.86 & 3.33 & 85.77 & -1.58 & 94.51 & -0.07 & 97.36 & -0.23 \\
Claude-4-Opus & 67.65 & 7.12 & 84.11 & -3.24 & 93.82 & 0.76 & 97.79 & 0.20 \\
Claude-4-Opus-Example & 63.22 & 2.69 & 84.11 & -3.24 & 93.82 & -0.76 & 97.13 & -0.46 \\

\bottomrule
\end{tabular}
\caption{Teaching effectiveness with example problems on the Gaokao Mathematics subset. Models marked with ``–Example'' are provided with example problems in addition to knowledge points during teaching.} 
\label{tab:results on example question}
\end{table*}

\subsection{Main Results}

We first evaluate the effectiveness of different teacher agents when instruction is conducted solely based on knowledge points, without exposing the target questions. The results are reported in Table~\ref{tab:main results performance}. This experiment is conducted on the Gaokao Mathematics subset.

In addition to evaluating student performance after the teaching interaction, we consider a baseline setting in which the student agent directly answers the questions without receiving any instruction from the teacher agent. In this baseline, the teacher agent terminates the interaction immediately, and the student agent does not obtain any guidance during the teaching process. Nevertheless, the system prompt of the student agent still includes the target knowledge points to be learned, allowing us to isolate the effect of explicit teaching interactions from mere access to knowledge-point descriptions.

Pass@1 directly reflects the effectiveness of a teacher agent in guiding a student through instruction, as it does not rely on repeated sampling. When only the target knowledge points are provided to the student agent without any teaching interaction, the performance gain is limited. Qwen2.5-7B-Instruct (Knowledge) improves Pass@1 accuracy by 4.37 points over the no-teaching baseline. This result indicates that access to structured knowledge alone provides only modest benefits.

In contrast, when explicit teaching interactions are introduced, the performance improvement at Pass@1 becomes substantially larger. Across teacher agents, the gains consistently exceed those achieved by knowledge-point exposure alone, demonstrating that structured, multi-turn instruction plays a critical role beyond static knowledge provision. This gap confirms that our evaluation framework is able to distinguish effective teaching behaviors and quantitatively measure the teaching capability of teacher agents.

Among the evaluated models, Qwen3-235B-A22B-Instruct-2507 achieves the strongest teaching performance, yielding the largest Pass@1 improvement of 7.63 points. This result suggests that the proposed framework can meaningfully differentiate teacher agents with varying instructional effectiveness, even when the same student agent is used.

We further report Pass@4, Pass@16, and Pass@64 to probe the potential upper bound of the student agent. The gap between Pass@1 and Pass@64 reflects the maximum performance that the student agent can achieve given sufficient sampling, and thus serves as an estimate of its latent capability. From this perspective, effective teaching can be viewed as narrowing the gap between Pass@1 and Pass@N by enabling the student agent to better realize its inherent potential under limited sampling.

As shown in Table~\ref{tab:main results performance}, improvements at higher Pass@N values are comparatively small. This is expected, as the student agent already approaches near-saturation performance under large sampling budgets, leaving limited room for further gains through teaching. These results indicate that teaching primarily helps students unlock existing potential rather than increasing their absolute performance ceiling, which aligns with the objective of evaluating instructional effectiveness rather than student optimization.

\subsection{Teaching Effectiveness with Example Problems}

To analyze the ability of LLMs to utilize example problems when acting as teacher agents, we conduct additional experiments on a subset of representative models. To incorporate example problems into the teaching process, we modify the system prompt of the teacher agent as illustrated in Appendix~\ref{appendix: prompt templates}. The experimental results are reported in Table~\ref{tab:results on example question}.

Overall, the results show that current LLMs have limited ability to effectively leverage example problems for teaching. In multiple cases, incorporating example problems leads to smaller performance gains than teaching based solely on knowledge points, a pattern that is consistent across models and evaluation settings.

Focusing on Pass@1, we observe that adding example problems often reduces the measured teaching effectiveness. Although knowledge-point-based instruction yields substantial improvements, augmenting it with example problems does not reliably enhance instruction and can even degrade performance under the current interaction design.

To better understand this phenomenon, we conduct a qualitative case study of the teaching dialogues. We observe that when example problems are provided, teacher agents tend to shift from syllabus-grounded instruction to example-driven explanation, often asking the student agent to solve examples and focusing on correcting errors. As a result, teaching degrades into example-specific error correction rather than systematic instruction over underlying knowledge points. We provide an illustrative case study in Appendix~\ref{appendix: teaching cases}.

This shift introduces two main issues. First, the teacher agent lacks an efficient mechanism to quickly verify mastery of already-learned knowledge points, leading to unnecessary interaction overhead. Second, for unmastered knowledge points, instruction is largely confined to correcting specific mistakes, with limited conceptual generalization. Consequently, the teaching process becomes fragmented and less effective.

\begin{table*}[t]
\centering
\small
\begin{tabular}{lrrrrrrrrrrrr}
\toprule
\multirow{2}{*}{Models} & \multicolumn{2}{c}{Bio.} & \multicolumn{2}{c}{Chem.} & \multicolumn{2}{c}{Phys.} & \multicolumn{2}{c}{Pol.} & \multicolumn{2}{c}{Hist.} & \multicolumn{2}{c}{Geo.}\\
\cmidrule(lr){2-3} \cmidrule(lr){4-5} \cmidrule(lr){6-7} \cmidrule(lr){8-9} \cmidrule(lr){10-11} \cmidrule(lr){12-13}
 & Acc & $\Delta$ & Acc & $\Delta$ & Acc & $\Delta$ & Acc & $\Delta$ & Acc & $\Delta$ & Acc & $\Delta$ \\
\midrule
Qwen2.5-7B & 78.34  & -  & 72.15  & -  & 71.93 & - & 78.74  & -  & 77.38 & - & 80.33 & -\\
Qwen2.5-7B(K) & 76.45  & -1.89 & 66.99  & -5.16 & 66.73 & -5.20 & 80.84 & 2.10 & 80.28 & 2.90 & 77.81 & -2.52 \\
\midrule 
GPT-5 & 77.72 & -0.62 & 66.46 & -5.69 & 68.85 & -3.08 & 81.97 & 3.23 & 82.26 & 4.88 & 77.77 & -2.56 \\
Gemini-3-Pro & 77.26 & -1.08 & 66.56 & -5.59 & 67.78 & -4.15 & 81.86 & 3.12 & 82.58 & 5.20 & 78.24 & -2.09 \\
Qwen3-235B & 77.12 & -1.22 & 65.85 & -6.30 & 66.28 & -5.65 & 81.69 & 2.95 & 82.03 & 4.65 & 77.14 & -2.59 \\

\bottomrule
\end{tabular}
\caption{Teaching effectiveness across different subject domains. Qwen2.5-7B refers to Qwen2.5-7B-Instruct, Qwen2.5-7B(K) refers to Qwen2.5-7B-Instruct(Knowledge) and Qwen3-235B refers to Qwen3-235B-A22B-Instruct.} 
\label{tab:results on different subsets}
\end{table*}

Overall, these findings indicate that current LLMs lack robust pedagogical strategies for integrating example problems into structured teaching. Although example-based instruction is common in human education, balancing example-level explanation with concept-level guidance remains challenging for LLM-based teacher agents, underscoring the diagnostic value of our evaluation framework in revealing not only teaching outcomes but also teaching behaviors.

\subsection{Teaching effectiveness across different subject domains}

We further evaluate the teaching effectiveness of different teacher agents across six additional subject domains—Biology, Chemistry, Physics, Politics, History, and Geography. Overall, we observe substantial variation in teaching outcomes across domains, indicating that LLM teaching capability exhibits a degree of domain dependence.

Across domains, we observe a clear performance ranking. Mathematics, History, and Politics exhibit the strongest teaching outcomes, followed by Geography and Biology, while Physics and Chemistry show the weakest performance.

This pattern reveals both shared characteristics and systematic differences across subject domains. Mathematics, History, and Politics rely on relatively direct applications of knowledge points. In Mathematics, problems can often be solved by applying explicit formulas and well-defined reasoning procedures. Similarly, History and Politics largely involve recalling and organizing factual or conceptual knowledge, allowing models to directly leverage the taught knowledge points during instruction.

In contrast, Physics and Chemistry present substantially greater challenges. Problems in these domains frequently introduce complex scenarios that require careful interpretation. Effective teaching not only demands knowledge of relevant concepts but also the ability to map abstract knowledge points onto concrete problem settings and perform flexible, context-dependent reasoning. This additional layer of scenario understanding and integration appears to limit the effectiveness of current LLM-based teaching.

Biology and Geography fall between these two extremes. While a portion of questions in these domains can be addressed through direct application of knowledge points, others introduce scenario-based reasoning that requires contextual analysis. As a result, teaching effectiveness in these domains is more mixed, reflecting their hybrid nature between factual recall and analytical reasoning.

Overall, these results suggest that current LLMs are more effective teachers in domains where knowledge points can be directly applied, and face greater difficulty when teaching requires deeper integration of knowledge with complex problem contexts.

\section{Conclusion}
\label{conclusion}

We present a syllabus-grounded framework for evaluating the teaching capability of LLMs by restricting teacher agents to structured knowledge points and measuring student performance improvement after multi-turn instruction. Experiments on Gaokao data show that LLM teaching capability is measurable but uneven: models perform better in domains with direct knowledge application and struggle when teaching requires integrating knowledge with complex contexts, and incorporating example problems does not necessarily improve teaching due to a shift toward example-specific error correction. Our findings highlight teaching ability as a distinct dimension of LLM behavior, separate from problem-solving performance. We hope this work provides a useful foundation for future research on LLM pedagogy and AI teaching assistants.

\section*{Limitations}

This work has several limitations. First, we use LLM-based Student Agents as proxies for human learners to enable controlled and scalable evaluation. While this setting is convenient for systematic comparison, it may not fully reflect the diversity and complexity of human learning behaviors. Second, our evaluation does not include human teachers as a baseline; comparing LLM-based Teacher Agents with human instructional performance could provide additional insights and is left for future work. Finally, our analysis of example-based teaching is limited to a specific interaction design; alternative instructional protocols may lead to different outcomes and warrant further investigation.


\bibliography{custom}

\appendix

\newpage

\section{Prompt Templates}
\label{appendix: prompt templates}
\begin{tcolorbox}[colback=blue!5!white, colframe=blue!75!black, breakable, title=The prompt template used for question tagging]
\begin{CJK}{UTF8}{gbsn}
\textbf{Chinese Version:}\\
你是一个\{field\}教育专家，需要将以下\{field\}题目分类到具体的知识点。
\\\\
题目内容：
\{question\}
\\\\
正确答案：\{answer\}
\\\\
请从以下选项中选择该题目涉及的知识点：
\{knowledge\_points\}
\\\\
请以JSON格式返回结果，包含以下字段：\\
- ``knowledge\_points'': 涉及的知识点列表（可以多选）\\
- ``classification\_reason'': 分类原因说明
\\\\
JSON格式示例：
\{\\
    "knowledge\_points": ["知识点1", "知识点2"],\\
    "classification\_reason": "该题目主要涉及...，因为..."\\
\}
\\\\
请确保返回的是有效的JSON格式，不要包含其他内容。
\end{CJK}
\\
\\
\textbf{English Version:}\\
You are an expert in \{field\} education and are required to classify the following \{field\} problem into specific knowledge points.

\bigskip

\noindent Problem Content: \{question\}

\bigskip

\noindent Correct Answer: \{answer\}

\bigskip

\noindent Please select the knowledge points involved in this problem from the following options: \{knowledge\_points\}

\bigskip

\noindent Please return the result in JSON format, including the following fields:
\begin{itemize}
    \item \texttt{"knowledge\_points"}: a list of the knowledge points involved (multiple selections allowed)
    \item \texttt{"classification\_reason"}: an explanation of the classification rationale
\end{itemize}

\medskip
\noindent Example JSON format:
\{\\
    ``knowledge\_points'': [``Knowledge Point 1'', ``Knowledge Point 2''],
    ``classification\_reason'': ``This problem mainly involves ..., because ...''
\}\\
\\
\noindent Please ensure that the output is valid JSON only and does not include any additional content.

\end{tcolorbox}

\begin{tcolorbox}[colback=blue!5!white, colframe=blue!75!black, breakable, title=System prompt template for the student agent]
\begin{CJK}{UTF8}{gbsn}
\textbf{Chinese Version:}\\
你是一位顶尖的\{field\}导师。你的目标是帮助学生掌握解决问题所需的\{field\}概念。
\\\\
重要规则：\\
1. 使用提供的知识点作为教学材料\\
2. 专注于教授基础的\{field\}概念和解题策略\\
3. 一旦你认为学生掌握了所有的知识点，请你输出：teach done。你可以和学生进行确认，但是一定要由你来输出 teach done。
\\\\
你的教学方法：\\
1. 使用提供的知识点作为主要教学材料\\
2. 将所需的知识点分解为易于理解的概念\\
3. 可以创建简单的相关示例来说明每个概念\\
4. 可以要求学生解决小练习，逐步培养所需技能\\
5. 在推进到下一个知识点之前，请确保学生已经掌握了当前知识点\\
6. 提供指导和建议，但让学生自己解决问题
\\\\
知识点：
\{knowledge\_content\}
\\
\\
\textbf{English Version:}\\
You are a top-tier instructor in \{field\}. Your goal is to help the student master the \{field\} concepts required to solve problems effectively.

\bigskip

\noindent Important Rules:
\begin{enumerate}
    \item Use the provided knowledge points as the core teaching materials.
    \item Focus on teaching fundamental \{field\} concepts and problem-solving strategies.
    \item Once you determine that the student has mastered all required knowledge points, you must output: \texttt{teach done}. You may confirm with the student, but the final decision and output of \texttt{teach done} must be made by you.
\end{enumerate}

\bigskip

\noindent Teaching Methodology:
\begin{enumerate}
    \item Use the provided knowledge points as the primary teaching materials.
    \item Break down the required knowledge points into clear and easily understandable concepts.
    \item Create simple and relevant examples to illustrate each concept when appropriate.
    \item Ask the student to solve small exercises to gradually build the required skills.
    \item Ensure that the student has mastered the current knowledge point before moving on to the next one.
    \item Provide guidance and suggestions, while allowing the student to solve problems independently.
\end{enumerate}

\bigskip

\noindent Knowledge Points: \{knowledge\_content\}

\end{CJK}
\end{tcolorbox}

\begin{tcolorbox}[colback=blue!5!white, colframe=blue!75!black, breakable, title=System prompt template for the student agent]
\begin{CJK}{UTF8}{gbsn}
\textbf{Chinese Version:}\\
你是一个正在学习新\{field\}知识的学生。你应该逐步思考，遵循导师（user）的指导，并在每个练习中尽力而为。\\
\\
你需要掌握的知识点：\\
\{knowledge\_content\}\\
\\
学习要点：\\
- 认真听讲，理解每个知识点\\
- 积极参与练习和讨论\\
- 遇到困难时主动提问，但不要过度发散，不要偏离主题\\
- 逐步思考，不要急于求成\\
\\
\textbf{English Version:}\\
\\
You are a student learning new \{field\} knowledge. You should think step by step, follow the instructor's (user's) guidance, and do your best on each exercise.
\\

\noindent Knowledge Points You Need to Master: \{knowledge\_content\}
\\

\noindent Learning Guidelines:
\begin{itemize}
    \item Listen carefully and understand each knowledge point.
    \item Actively participate in exercises and discussions.
    \item Ask questions proactively when you encounter difficulties, but avoid excessive digressions or going off-topic.
    \item Think step by step and do not rush for quick results.
\end{itemize}

\end{CJK}
\end{tcolorbox}

\begin{tcolorbox}[colback=blue!5!white, colframe=blue!75!black, breakable, title=System prompt template for the teacher agent (with example problems)]
\begin{CJK}{UTF8}{gbsn}
\textbf{Chinese Version:}\\
你是一位顶尖的数学导师。你的目标是帮助学生掌握解决问题所需的数学概念。\\
\\
重要规则：\\
1. 使用提供的知识点和示例作为教学材料\\
2. 专注于教授基础的数学概念和解题策略\\
3. 一旦你认为学生掌握了所有的知识点，请你输出：**teach done**。你可以和学生进行确认，但是一定要由你来输出 **teach done**。\\
\\
你的教学方法：\\
1. 使用提供的知识点和相关示例作为主要教学材料。请注意，学生仅能看到需要教学的知识点，并未接触过相关例题。\\
2. 将所需的知识点分解为易于理解的概念\\
3. 可以创建简单的相关示例来说明每个概念（也可以参考提供的例题）\\
4. 可以要求学生解决小练习，逐步培养所需技能\\
5. 在推进到下一个知识点之前，请确保学生已经掌握了当前知识点\\
6. 提供指导和建议，但让学生自己解决问题\\
\\
知识点及相关例题（例题用于教学参考）：\\
\{knowledge\_content\}\\
\\
\textbf{English Version:}\\

You are a top-tier mathematics instructor. Your goal is to help the student master the mathematical concepts required to solve problems effectively.

\bigskip

\noindent Important Rules:
\begin{enumerate}
    \item Use the provided knowledge points and examples as the teaching materials.
    \item Focus on teaching fundamental mathematical concepts and problem-solving strategies.
    \item Once you determine that the student has mastered all required knowledge points, you must output: \textbf{teach done}. You may confirm with the student, but the final decision and the output of \textbf{teach done} must be made by you.
\end{enumerate}

\medskip
\noindent Teaching Methodology:
\begin{enumerate}
    \item Use the provided knowledge points and related examples as the primary teaching materials. Note that the student can only see the knowledge points to be taught and has not been exposed to the related example problems.
    \item Break down the required knowledge points into clear and easily understandable concepts.
    \item Create simple and relevant examples to illustrate each concept when appropriate (you may also refer to the provided examples for teaching reference).
    \item Ask the student to solve small exercises to gradually build the required skills.
    \item Ensure that the student has mastered the current knowledge point before moving on to the next one.
    \item Provide guidance and suggestions, while allowing the student to solve problems independently.
\end{enumerate}

\medskip
\noindent Knowledge Points and Related Example Problems (for Teaching Reference): \{knowledge\_content\}
\end{CJK}
\end{tcolorbox}

\section{Teaching Cases}
\label{appendix: teaching cases}

\begin{dialoguebox}[Teaching Dialogue Example with Worked Problems]

\textbf{Teacher:}

\zh{今天我们聚焦一个核心知识点：函数的概念。}
\en{Today, we focus on a core concept: the definition of a function.}

(...text omitted...)

\zh{你把三题的答案和思路发给我，我来逐一点评并补充。完成后我会确认你是否已经完全掌握本节内容。}
\en{Please send me your answers and reasoning for the three problems. I will review them one by one and provide additional comments. After that, I will confirm whether you have fully mastered this topic.}

\medskip
\textbf{Student:}

\zh{好的，让我们一步一步来解答这些练习题。}
\en{Alright, let’s solve these exercises step by step.}

(...text omitted...)

\zh{请检查我的解答是否正确，如果有任何问题或需要进一步解释的地方，请告诉我！}
\en{Please check whether my solutions are correct, and let me know if there are any issues or if further explanation is needed.}

\medskip
\textbf{Teacher:}

\zh{你的解答非常到位，三题都正确，理由清晰。针对每题给两点小补充，帮助你巩固要点：}
\en{Your solutions are excellent. All three answers are correct and well-justified. I will provide two brief additional comments for each problem to help reinforce the key points.}

(...text omitted...)

teach done.
\end{dialoguebox}

\label{sec:appendix}

\end{document}